\pdfoutput=1

\documentclass[nohyperref]{article}

\usepackage[accepted]{icml2022}

\usepackage{microtype}
\usepackage{graphicx}
\usepackage{subfigure}
\usepackage{booktabs} 
\usepackage{hyperref}

\usepackage{icml2022}

\usepackage{times}
\usepackage{latexsym}

\usepackage{natbib}

\usepackage{xcolor}
\usepackage{todonotes}
\usepackage{tabularx} 

\newcommand{\dontcite}[1]{} 
\newcommand{\hide}[1]{} 

\usepackage[T1]{fontenc}

\usepackage[utf8]{inputenc}

\begin{document}

\twocolumn[
\icmltitle{Recognizing and Extracting Cybersecurity Entities from Text}

\begin{icmlauthorlist}
\icmlauthor{Casey Hanks}{sch}
\icmlauthor{Michael Maiden}{sch}
\icmlauthor{Priyanka Ranade}{sch}
\icmlauthor{Tim Finin}{sch}
\icmlauthor{Anupam Joshi}{sch}
\end{icmlauthorlist}

\icmlaffiliation{sch}{University of Maryland, Baltimore County, Baltimore, MD 21250}

\icmlcorrespondingauthor{Casey Hanks}{chanks1@umbc.edu}
\icmlcorrespondingauthor{Michael Maiden}{mmaiden1@umbc.edu}

\icmlkeywords{Cybersecurity, Natural Language Processing, NLP, information extraction}

\vskip 0.3in
]

\printAffiliationsAndNotice{}

\begin{abstract}
Cyber Threat Intelligence (CTI) is information describing threat vectors, vulnerabilities, and attacks and is often used as training data for AI-based cyber defense systems such as Cybersecurity Knowledge Graphs (CKG). There is a strong need to develop community-accessible datasets to train existing AI-based cybersecurity pipelines to efficiently and accurately extract meaningful insights from CTI. We have created an initial unstructured CTI corpus from a variety of open sources 
that we are using to train and test cybersecurity entity models using the spaCy framework and exploring self-learning methods to automatically recognize cybersecurity entities. We also describe methods to apply cybersecurity domain entity linking with existing world knowledge from Wikidata. Our future work will survey and test spaCy NLP tools, and create methods for continuous integration of new information extracted from text.
\end{abstract}

\section{Introduction}

Cyber Threat Intelligence (CTI) is data that has been analyzed to attempt to uncover the mechanics of and purpose for a cyber-attack. CTI is vital to cybersecurity professionals for staying up to date on information about new attacks, malware, and the actions of various threat actors. Professionals use this information in a variety of use cases, including campaign tracking, threat monitoring, and actor profiling. However, with a vast amount of CTI being released and updated every day, it is increasingly challenging for human analysts to track the data efficiently and effectively. It is important to develop methods for automated cybersecurity knowledge extraction, to consolidate CTI insights and make it easier for experts to access and use.

Named Entity recognition (NER) is a critical component of automated knowledge extraction, allowing natural language models to label instances of real-world entities in text. To accomplish this, Natural Language Processing (NLP) models must be trained on very large corpora of human-annotated text. There are a number of domain-agnostic text corpora available for training models on generic entity types such as \textit{Person, Organization}, and \textit{Date}. However, general domain entity types are not sufficient for more specialized fields like cybersecurity because they are unable to recognize cybersecurity-specific entities such as malware-type, operating system, or attack-type. These are necessary for downstream tasks like malware analysis, attack and vulnerability classification, and building cybersecurity knowledge graphs  \cite{mulwad11,ajoshi13,gao2021review, Georgescu21}. Unlike fields like medicine and law, cybersecurity has few comprehensive training datasets that are available \textit{and} continuously updated.

\begin{figure}[t]
    \centering
    \fbox{\includegraphics[width=0.95\columnwidth]{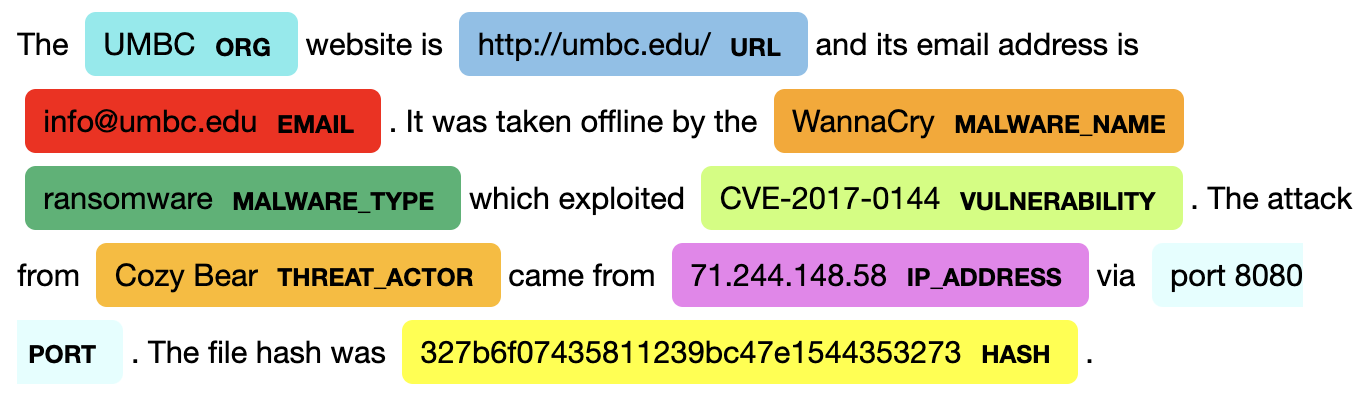}}
    \vspace{-1mm} 
    \caption{Text showing additional \hide{domain-specific} entity types found.}
    \label{fig:displacy}
    \vspace{-4mm} 
\end{figure}
Our goal is to extend existing entity recognition datasets for the cybersecurity domain \cite{alam22, bridges} to support a general module we call \textit{CyEnts} that can be used in systems that extract cybersecurity information from text. Current public datasets do not recognize entities such as \textit{threat actors, campaigns, or malware}, imperative for training machine learning systems that aim to understand fine-grained threat actor tactics, techniques, and procedures (TTPs). The kinds of entities we need to identify are also more general and diverse than the "named entities" sought by current NLP systems designed to extract information from general domains.

Cyber analysts typically rely on both structured sources like Common Vulnerabilities and Exposures (CVE) records, as well as unstructured, free-text blogs like security vendor blogs to obtain actionable intelligence. Most of the existing entity recognition datasets for the cybersecurity domain target structured data, as the entities are formally defined, providing concrete annotation mappings for entity types. Our research focuses on extending entity recognition capabilities for unstructured sources, since these typically provide the most recent information for analysts. Before CTI is concatenated into structured formats like CVE records, the information is first released through online media.
 
To create a diverse dataset for malware analysis, we have obtained articles, blog posts, and vendor reports from eight different sources, covering high level topics such as adversary motives and origination, to technical analysis of malware behaviors and campaigns. We develop a human annotation study to create high-quality training datasets for machine learning based entity recognition models, using the spaCy framework. Table~\ref{tab:accents} shows the current list of cybersecurity entity types we support in addition to spaCy's OntoNotes types \cite{ontonotes2007}. We take a \textit{data-centric} approach when creating the training dataset, by using intensive evaluation criteria for entity recognition annotations.

\begin{table}[t]
\vspace{-3mm}
\caption{New cybersecurity-relevant entity types complement the Ontonotes types supported by spaCy and other NLP systems.}
\vspace{1mm}
\centering
\fbox{%
\begin{tabular}{l | l}
Malware\_Name & Campaign \\
Malware\_Type & IP\_Address \\
Software\_Name& Protocol \\ 
Version\_Tag & Threat\_Actor \\ 
Vulnerability & Operating\_System \\
Attack\_Type & Hash \\
Programming\_Language & URL \\
Email & Path \\
File\_Extension & Function\\
CVE & Port\\
\end{tabular} }
\vspace{-4mm} 
\label{tab:accents}
\end{table}

\section{Related Work}

\subsection{AI-Based Cybersecurity}

The variety and growth in cybersecurity exploits, vulnerabilities, and threat actors has encouraged the integration of AI-based cyber defense systems for safeguarding critical systems \cite{wirkuttis2017artificial}. The goal of these systems is to provide actionable and relevant insights to analysts that require the information for immediate operation. This is becoming especially critical, as the diversity and amount of Cyber Threat Intelligence (CTI) is rapidly expanding \cite{tounsi2018survey}. CTI is often \textit{open sourced} and can include data sources such as application logs, malware binaries, network traffic data, and unstructured and semi-structured text. This data is  collectively shared across multiple vendors, researchers, and cybersecurity professionals for enhanced situational awareness and intrusion detection and prevention.  Examples of threat sharing platforms include MISP \cite{wagner2016misp}, STIX \cite{stix2}, and TAXII \cite{connolly2014trusted}.

In this paper, we focus on leveraging textual cybersecurity data, which can be commonly found across security blogs, the dark web, and social media. There have been several systems that have been developed that transform free-text cybersecurity into more structured formats for AI-based cyber defense system usage \cite{samtani2020proactively, cybertwitter, arnold2019dark, ranade2021cybert}. In particular, several Cybersecurity Knowledge Graphs (CKGs) have been developed to represent disparate CTI data in machine-readable formats, and are used as training data for machine learning systems \cite{mittal2019cyber, pinglerelext,piplai2020creating}.

\subsection{Knowledge Representation}

Ontologies are one of the primary building blocks of the Semantic Web \cite{berners2001semantic}. When populating Ontologies with real world data, information can be associated together and reasoned over through the \textit{web of linked data} \cite{berners2001semantic}. Examples of popular knowledge graphs include DBPedia \dontcite{lehmann2015dbpedia} and Wikidata \cite{wikidata2014}. 
Machine learning and semantic technologies are more recently, jointly used together for tasks such as language modeling \cite{agarwal2020knowledge}, question-answering \cite{wang2016deep}, and information retrieval \cite{wise2020covid}.

\section{System Overview}

\begin{figure}[t]
    \centering
    \fbox{\includegraphics[width=\columnwidth]{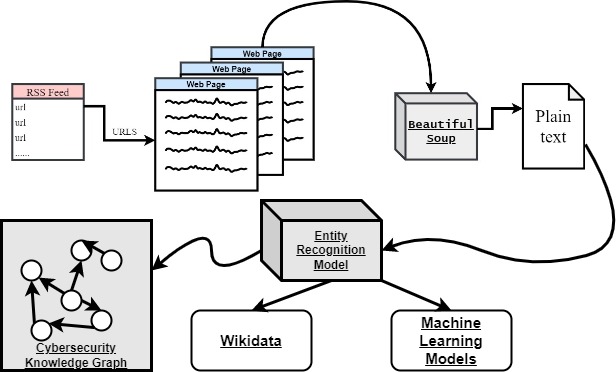}}
    \caption{Diagram of the proposed complete system}
    \label{fig:system}
    \vspace{-4mm} 
\end{figure}

CyEnts will incorporate a continuous, up to date corpus of ready to train entity recognition datasets, achieved through a combination of web scraping and entity recognition techniques. This data can then be used to form triples to populate a knowledge graph, train other machine learning systems, as well as link to Wikidata knowledge to provide extensibility to other interoperable systems. In this section we discuss more of how our system proposes to accomplish this.

\subsection{Web Scraping}

We use a combination of Python \textit{Requests} and \textit{BeautifulSoup} libraries to scrape web articles that contain CTI from cybersecurity vendor reports/blogs. We regularly update this corpus through periodic scraping to incorporate new articles, preventing stale training data. Our growing corpus can be considered a gold standard dataset for the cybersecurity community due to two primary factors: the source reputability and their breadth of CTI examples. The first is the popularity of the sources utilized across the greater cybersecurity community. Vendor reports and blogs produced by organizations such as \textit{McAfee, Mandiant, FireEye}, and \textit{Juniper Networks}, describe up-to-date and pertinent vulnerability, attack, and adversary information to aid cybersecurity professionals in mitigating incoming attacks, quickly addressing open vulnerabilities, and thwarting future threats \cite{samtani2020cybersecurity}. Secondly, vendor reports and blogs do not follow a standard format and are therefore inherently diverse in their communication of vulnerabilities, exploits, and threats, providing the system with multiple perspectives and examples.

We collect our data through RSS feeds, which list a number of articles on each site, in order of publication. This text is stored with no newline characters to simplify tasking for the human annotations, described in Section \ref{section:annotations}. We use spaCy facilities to split the text into sentences, and then group the sentences by topic to produce ``paragraphs'' using NLTK's textiling tools \cite{loper2002nltk}. We currently have about 25,000 sentences from over 380 text articles\footnote{Our textual data is available at \url{https://github.com/UMBC-Onramp/CyEnts-Cyber-Blog-Dataset}}

\subsection{Entity Recognition}

The first step in creating an entity recognition system is to define the entity types to recognize. We have created a set of entity types, listed in Table~\ref{tab:accents}, after extensive literature review and multiple revisions. We include common entity types such as \textit{Software\_Name} and \textit{Version\_Tag}, as well as more fine-grained types, like \textit{Threat\_Actor} and \textit{Malware\_Type}.  We believe this set of entity types to be gold standard for malware analysis, due to the inclusion of descriptive types for entities imperative for processing and understanding malicious activity and file behavior. For example, the operating system a malware targets, or the threat actor involved, can provide additional real-world context to a malware analysis task. We have created a spaCy pipeline for recognizing these entity types through use of human annotations of text, and rule based methods, detailed below.

\subsubsection{Human Annotations}
\label{section:annotations}

To create ground-truth annotations, we task a group of six human annotators to recognize entities in the corpus data. The annotators have backgrounds in Computer Science and Cybersecurity. These annotations, along with the rule-based methods described in Section \ref{section:rulebased}, will later be used to train machine-learning based methods.

\begin{figure}[t]
    \centering
    \includegraphics[width=\columnwidth]{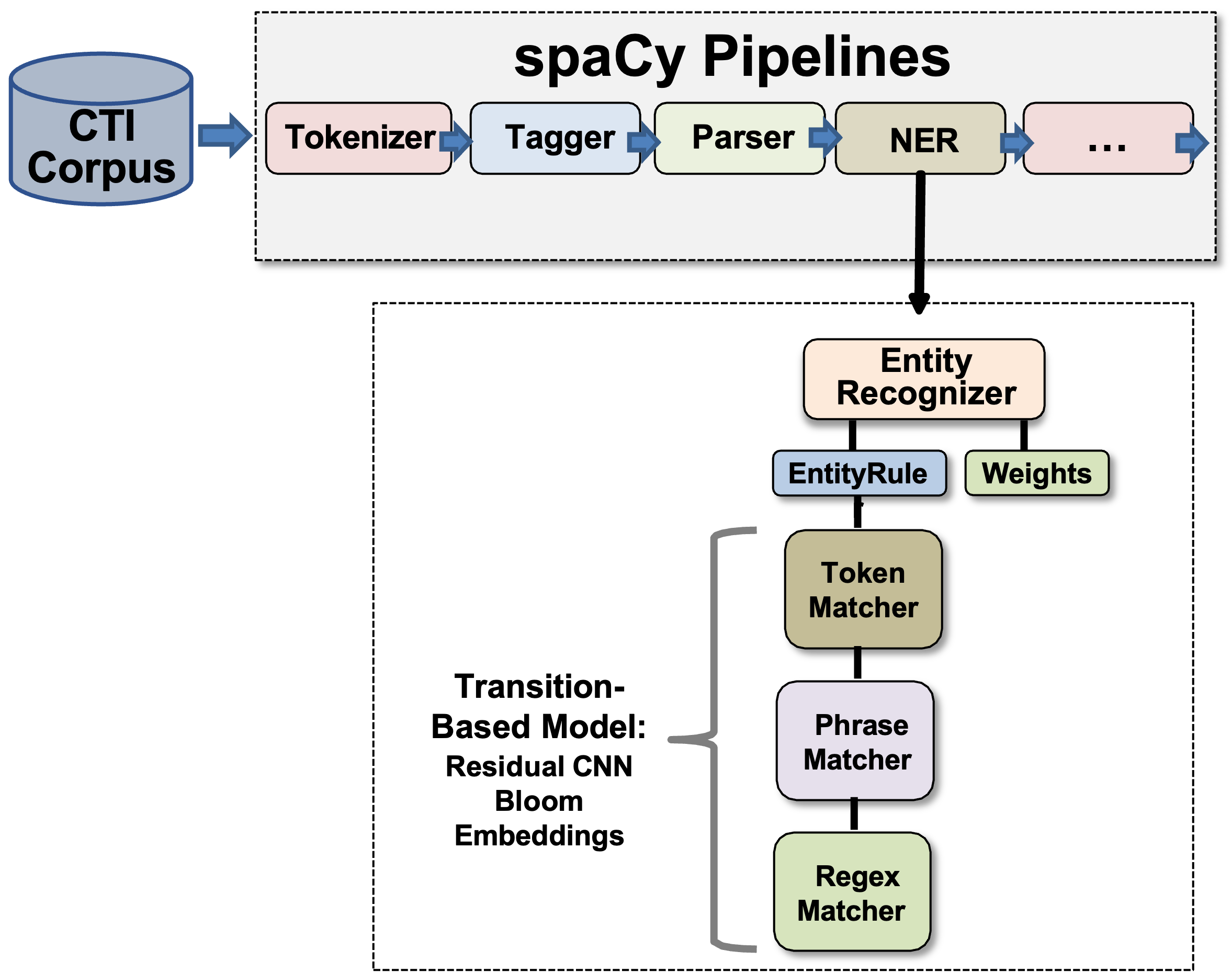}
    \caption{Components of the entity recognition pipeline}
    \label{fig:ERpipeline}
\end{figure}

To ensure quality annotations, each annotator received training, documentation detailing the definition of each entity type, and multiple in-context examples. We divided the human annotators into three groups. Each annotator within a group was tasked to annotate the same text independently. Once complete, we calculate the Inter Annotator Agreement, which means that only entities that both annotators agreed upon would be accepted into the final dataset. To conduct our annotation task, we use the \textit{Prodigy} annotation tool. This tool allows for seamless integration with the spaCy NLP framework. The resulting annotated compiled dataset is used to train and test an \textit{entity recognizer} model. The architecture of the entity recognizer model we utilized is displayed in Figure \ref{fig:ERpipeline}, and is further explained in the following section. In the first round of annotations each annotator was given the first ten paragraphs of ten randomly selected articles, for a combined total of 1339 annotated sentences. The results from the first round are provided in Section \ref{section:eval}.

 
\subsubsection{Rule Based Methods}
\label{section:rulebased}

We utilize \textit{spaCy pipelines}, which are several in-built models and libraries utilized for different Natural Language Processing (NLP) tasks. The available pipelines and the specific models we implement are shown in Figure \ref{fig:ERpipeline}. The Entity Recognizer model part of the NER pipeline is utilized for our initial task of extracting and recognizing entities in unstructured CTI. The model employs a transition-based algorithm which utilizes a combination of \textit{Bloom Embeddings} and a Residual Convolutional Neural Network (CNN) \cite{lample2016neural}. More specifically, spaCy employs the \textit{Embed, Encode, Attend, Predict} framework \cite{honnibal2016embed}. Each token is first embedded using a \textit{Bloom filter}, which is a hashed embedding dictionary where word hashes, rather than the actual words, are set as keys.The words are then encoded into a sentence matrix with a Residual CNN. The sentence matrix is further reduced into a single vector to be used in a feed-forward neural network for prediction. This necessarily allows us to lose irrelevant information and attend to the most useful features. We can then use the standard feed-forward neural network for inference, to predict the target representation (entity label). 

There are many entity types that can be recognized using rule-based methods. The types fall into two major categories, those that have a finite number of relevant examples and ones with a regular format. For the former category of entities we can use \textit{Gazetteers} \cite{SOng2020} of relevant entities of a particular type. We have developed Gazetteers for the entity types, Operating\_System, File\_Extension, Attack\_Type, Programming\_Language, Malware\_Type, and Protocol. The lists for the Gazetteers were created by querying Wikidata to find the most prominent instances (direct or inherited) of all of the types, to make sure that we include entities that are relevant. More information on Wikidata entity linking can be found in Section \ref{section:wikidata}. These lists were then manually reviewed to ensure they were encompassing.

The second category where rule-based methods are helpful are entities which have a strict format, such as IP address, that can be recognized by regular expressions. We use regular expressions with IP\_Address, Hash, Port, and CVE vulnerabilities (alongside regular entity recognition for named vulnerabilities). We also use spaCy's built-in rule based recognition for Email addresses and URLs. We included these recognizers in the pipeline used to prime the second round annotations, so that the process was more streamlined.

\subsubsection{Wikidata Entity Linking}
\label{section:wikidata}

Where possible, we link entity mentions to items in Wikidata \cite{wikidata2014} which has about one billion facts on about 100 million items. It has a Web interface to support exploration and editing by people, a set of APIs to access its information programmatically, and a SPARQL \dontcite{sparql} endpoint for querying its RDF knowledge graph. Wikidata's ontology has a very fine-grained type system with more than two million types and over 10,000 properties.  The properties of Wikidata items we primarily use are its types, super-types, label, aliases, and description. Many other Wikidata properties are available for future use.

Applying entity linking to a domain involves being able to recognize both the Wikidata items and properties that are very relevant to the domain as well as those that are likely to be irrelevant. We identified custom sets of Wikidata types and properties to support the cybersecurity domain for this project. Given an entity mention, it uses Wikipedia's existing search APIs to retrieve an initial set of candidate entities, typically between 20 and 50.  The API searches against the text in labels, aliases, and description as well some property values.  

The results are then ranked using a variety of matching features, item prominence, and the semantic similarity of the entity's text context compared to its Wikidata description and initial sentence in its DBpedia \cite{lehmann2015dbpedia} abstract.
This approach can help CyEnts successfully assign the type Threat\_Actor to \textit{Lazarus} in the sentence 
\begin{quote}
"Lazarus was behind the WannaCry attack"
\end{quote}
and also link it to the North Korean hacker organization \textit{Lazarus Group} (Q19284445)
even though Lazarus matches all or part the names of more than 1,000 Wikidata items.

\subsection{Evaluation}
\label{section:eval}

The initial Precision, Recall, and F-score for the first round of the NER dataset is provided in Table \ref{Scores}. The breakdown of the Precision, Recall, and F-score for each entity type is listed in Table \ref{EntityScores}. We have omitted spaCy general types as well as cybersecurity classes with low representation from the table, due to lack of examples present for evaluation. We discuss increasing the representation of sparse labels and tuning the human evaluation study to produce better annotations for training more robust models in Section \ref{section:improvingannotations}.

Unlike entity recognition for more structured formats like CVE records, blogs contain a greater degree of variety in sentence structure, information, and style, adding increased difficulty for recognizing relevant entities, in agreeable and universal formats. We are improving our first round of annotations by employing rule-based strategies, in addition to aiding the annotators with further training.  We discuss other lessons learned from our first round below.

\begin{table}[t]
\vspace{-3mm}
\caption{Breakdown of number of annotations for each Entity Type}
\label{NumAnnotations}
\vskip 0.15in
\centering
\small
\setlength\tabcolsep{2pt}
\begin{center}
\begin{small}
\begin{sc}
\begin{tabular}{lc|ccr}
\toprule
Type & Number & Type & Number \\
\midrule
Version\_Tag & 17 & EVENT & 1\\
Malware\_Type & 33& GPE & 36\\
Threat\_Actor & 24& LAW & 0\\
Vulnerability & 26& MONEY & 3\\
Filename & 37& ORG & 149\\
Protocol & 29 & PRODUCT & 1\\
Port & 1 & TIME & 6\\
Software\_Name & 53 & DATE & 148 \\
Malware\_Name & 118 & FAC & 0 \\
Tool & 1 & LANGUAGE & 4 \\
Campaign & 14 & LOC & 0 \\
Operating\_System & 35 & NORP & 10  \\
Filepath & 7 & QUANTITY & 0\\
Process & 10 & PERSON & 16 \\
Attack\_Type & 2\\
\bottomrule
\end{tabular}
\end{sc}
\end{small}
\end{center}
\vskip -0.25in
\end{table}

\begin{table}[t]
\caption{Precision, recall, and F-score of our initial model}
\label{Scores}
\vskip 0.15in
\centering
\small
\setlength\tabcolsep{2pt}
\begin{center}
\begin{small}
\begin{sc}
\begin{tabular}{lcr}
\toprule
Precision & Recall & F-score \\
\midrule
70.77 & 60.53 & 65.25 \\
\bottomrule
\end{tabular}
\end{sc}
\end{small}
\end{center}
\vskip -0.25in
\end{table}

\begin{table}[t]
\caption{Precision, Recall, and F-score for each entity type in our initial model}
\label{EntityScores}
\vskip 0.15in
\centering
\small
\setlength\tabcolsep{2pt}
\begin{center}
\begin{small}
\begin{sc}
\begin{tabular}{lccr}
\toprule
Entity Type & Precision & Recall & F-Score \\
\midrule
Filename & 50.00 & 40.00 & 44.44 \\
Malware\_Name & 60.00 & 84.00 & 70.00 \\
Vulnerability & 57.14 & 100.00 & 72.73 \\
Operating\_System & 71.43 & 71.43 & 71.43 \\
Software\_Name & 90.00 & 69.23 & 78.26 \\
Version\_Tag & 25.00 & 33.33 & 28.57 \\
Filepath & 0.00 & 0.00 & 0.00 \\
Protocol & 33.33 & 10.00 & 15.48 \\
Threat\_Actor & 100.00 & 100.00 & 100.00 \\
Campaign & 50.00 & 33.33 & 40.00 \\
Malware\_Type & 0.00 & 0.00 & 0.00 \\
\bottomrule
\end{tabular}
\end{sc}
\end{small}
\end{center}
\vskip -0.25in
\end{table}

\subsubsection{Representation of Entity Types}

Table~\ref{NumAnnotations} displays the number of annotations for each entity type annotated during the first round of annotations. The left half of the table includes cybersecurity domain entity types that we defined. The right half of the table includes general spaCy types that the out of the box model is trained to recognize. The general types are pre-populated during the annotation engine instantiation. We modified this provided for the second round of annotations. More information on modifications to this list is described in Section \ref{section:improvingannotations}.

We observe that the occurrence of entity types in a document can impact the number of annotations for certain types, unbalancing the data distribution when training. For example, we noticed limited occurrence of spaCy general types in the CTI corpus. However, certain general types like \textit{ORG} and \textit{DATE} have higher frequency. In terms of cybersecurity specific types, we notice that \textit{Malware\_Name} has high frequency, as most of the articles describe technical details of APT groups and malware types. Since some of the less frequent types did not have enough annotations to train the Entity Recognizer model, the F score lowered as a result.


\subsubsection{Inter-Annotator Agreement}

In addition, there was high disagreement between annotators. Taking the maximum of each group's annotations, there was a total number of 1755 annotations. However only 781 annotations were accepted, based on the Inter-Annotator Agreement scores. One observation to explain the disagreement is multiple definitions of entity types, such as \textit{Product} and \textit{Software Name}. Broadly, there are many variations for the methods in which one could annotate these entities. For example, the annotators often defined software libraries as products. In addition, discrepancies such as \textit{Windows} versus \textit{Windows OS}, can drastically change the distribution of the data.  A similar conflation happened between \textit{Software\_Name} and \textit{Tool}. Another issue was in entities that could be classified as multiple types. Consider the following example: ``Microsoft Word''. An annotator could recognize is Microsoft an \textit{ORG} and Word the \textit{Software\_name}, or could merge the terms together as the name of a single software type. Lastly, the final potential issue was syntactical: if a process was called \textit{RunGame()},  some annotators would include the parentheses and some not. The annotator disagreement issues also contributed to some entity types lacking annotations in the final dataset.

\subsubsection{Improving Human Annotation Study}
\label{section:improvingannotations}

To improve our dataset we are undergoing a second round of annotations, with comprehensive changes such as the introduction of the rule-based entity pre-population, and further annotator training. Another change we made was modifying the initial list of entity types. For example, we removed the type \textit{Tool} because it was difficult to differentiate it from \textit{Software}, and in many cases the types were semantically, the same. We deemed \textit{Process} to be more general and nebulous to annotate well, so we redefined it into \textit{Function} as it was more agreeable to multiple sub domain definitions. We also redefined \textit{Filepath} to just \textit{Path}, and added the \textit{File\_Extension} type. To further help mitigate annotator disagreement, we provided more in-depth training and documentation to the annotators.

Our strategy for mitigating the lack of annotations for certain types is to implement rule-based recognition for these types so they require fewer annotations to recognize at an acceptable level. We especially target entities with low annotation numbers for this, such as \textit{Attack\_Type} or \textit{Port}. These rule-based recognizers also correspond to entities that had very low F-scores in \ref{EntityScores}. Entity types such as \textit{Filepath}, \textit{Malware\_type}, \textit{Protocol}, and others which had low F-scores are those that we are implementing rule-based recognizers for in the updated pipeline, whereas entity types such as \textit{Malware\_Name}, \textit{Software\_Name}, and others perform fairly well without these and we believe will perform even better with more annotations.

These rule-based recognizers were also applied during the second round of annotation, which also helps reduce annotator confusion as the entities are already labeled. Our final measure was to increase the size of the annotation set to 150 paragraphs of text for each annotator to ensure each entity receives a consistent set of annotations.

\subsection{Conclusion}

We described a preliminary framework for improving entity recognition and linking for the cybersecurity domain. We created a corpus of cyber threat intelligence, developed a human annotator study to create high quality and shareable cybersecurity entity recognition datasets, tested spaCy framework capabilities to improve automatic entity extraction, and identified custom sets of Wikidata types and properties to support the cybersecurity domain.

In future work, we will integrate our Wikidata entity linker into the pipeline, add a \textit{coreference module} to link pronominal and nominal mentions to entities, train a relation-extraction module for the domain, and adapt our earlier work to extract information on cybersecurity events \cite{satyapanich20}. We also plan to develop a system to automatically do periodic web scraping and processing of the new text. The cybersecurity entities, relations, and events will be used to populate and extend existing CKGs \cite{piplai2020creating,mitra22}.

\section*{Acknowledgements}
This research was supported by grants from NSA and the National Science Foundation (No. 2114892).

\nocite{honnibal2020spacy} 
\bibliographystyle{icml2022}
\bibliography{icml22}
\end{document}